\documentclass[letterpaper]{article} 
\usepackage[]{aaai23}  
\usepackage{times}  
\usepackage{helvet}  
\usepackage{courier}  
\usepackage[hyphens]{url}  
\usepackage{graphicx} 
\usepackage{natbib}  
\usepackage{caption} 
\usepackage{microtype}
\usepackage{subfigure}
\usepackage{booktabs} 

\usepackage{algorithm}
\usepackage{algorithmic}

\usepackage{amsmath}
\usepackage{amssymb}
\usepackage{mathtools}
\usepackage{amsthm}

\usepackage{tabularx}
\usepackage{booktabs}
\usepackage{animate}

\renewcommand{\vec}[1]{\mathbf{#1}}
\renewcommand{\Re}{\mathbb{R}}

\DeclareMathOperator{\fk}{\textsc{FK}}
\DeclareMathOperator{\att}{\textsc{ATT}}
\DeclareMathOperator{\mha}{\textsc{MHA}}
\DeclareMathOperator{\mlp}{\textsc{MLP}}
\DeclareMathOperator{\relu}{\textsc{ReLu}}
\DeclareMathOperator{\layernorm}{\textsc{LayerNorm}}
\DeclareMathOperator{\softmax}{\textsc{softmax}}
\DeclareMathOperator{\concat}{\textsc{concat}}

\usepackage[dvipsnames]{xcolor}

\usepackage{pifont}
\newcommand{\ballot}{\ding{55}}
\renewcommand{\checkmark}{\ding{51}}

\usepackage[capitalize,noabbrev]{cleveref}

\usepackage{newfloat}
\usepackage{listings}
\DeclareCaptionStyle{ruled}{labelfont=normalfont,labelsep=colon,strut=off} 
\floatstyle{ruled}
\newfloat{listing}{tb}{lst}{}
\floatname{listing}{Listing}
\title{Motion In-betweening via Deep $\Delta$-Interpolator}

\title{Motion In-betweening via Deep $\Delta$-Interpolator}
\author {
    Boris N. Oreshkin\textsuperscript{* \rm 1},
    Antonios Valkanas\textsuperscript{* \rm 2}, 
    Félix G. Harvey\textsuperscript{\rm 1}, 
    Louis-Simon Ménard\textsuperscript{\rm 1}, 
    Florent Bocquelet\textsuperscript{\rm 1}, 
    Mark J. Coates\textsuperscript{\rm 2}, 
}
\affiliations {
    \textsuperscript{*}Equal contribution, \textsuperscript{\rm 1}Unity Technologies, \textsuperscript{\rm 2}McGill University\\
    boris.oreshkin@gmail.com, antonios.valkanas@mail.mcgill.ca 
}

\begin{document}

\maketitle
\begin{abstract}
We show that the task of synthesizing human motion conditioned on a set of key frames can be solved more accurately and effectively if a deep learning based interpolator operates in the delta mode using the spherical linear interpolator as a baseline. We empirically demonstrate the strength of our approach on publicly available datasets achieving state-of-the-art performance. We further generalize these results by showing that the $\Delta$-regime is viable with respect to the reference of the last known frame (also known as the zero-velocity model). This supports the more general conclusion that operating in the reference frame local to input frames is more accurate and robust than in the global (world) reference frame advocated in previous work. Our code is publicly available at
\url{https://github.com/boreshkinai/delta-interpolator}.
\end{abstract}

\section{Introduction}

Computer generated imagery (CGI) has enabled the generation of high quality special effects in the movie industry with the movie making process becoming ever more technologically intensive as time progresses~\citep{ji2010production}. Animating 3D characters is integral to the process of creating high quality CGI in films and in a variety of other contexts. Unsurprisingly, 3D computer animation is at the core of modern film, television and video gaming industries~\cite{beane20123d}. Traditional animation workflows rely on the 3D sketches of the most important \emph{key-frames} in the produced character motion sequence drawn by experienced artists. For example, one viable animation workflow relies on a senior artist drawing a few 3D key-frames pinpointing the most important aspects of the authored 3D sequence and a few less senior animators working out the details in-between these key-frames. Another viable computer aided animation workflow relies on massive motion capture (MOCAP) databases, enabling an animator to query suitable motion sequences from them based on a variety of inputs, such as a few key-frames, root-motion curve, meta-data tags, etc. The discovered motion fragments can be further blended to create new motion sequences with desired characteristics. Time and effort involved in exploring and traversing such diverse databases to find and blend suitable motion segments can be a significant bottleneck.

\begin{figure}[ht]
    \centering
    \animategraphics[loop,autoplay,controls,width=\linewidth]{24}{animation1/GT_INT_DT_run1_frame236_int30_distance-}{0}{52}
    \caption{The demonstration of the proposed $\Delta$-interpolator approach (right, green), Linear interpolator baseline (middle, yellow), ground truth motion (left, white). Positional errors are indicated using red mask. \textbf{The animation is best viewed in Adobe Reader.}}
    \label{fig:animation_demo}
\end{figure}

These tedious processes can be automated to a certain extent by filling the missing in-between frames using an array of techniques including linear interpolation, motion blending trees and character specific MOCAP libraries. However, a more promising avenue to enable data-driven motion generation has been unblocked more recently by the advancement in deep learning algorithms~\cite{harvey2020robust,duan2021singleshot}. Deep learning models that are trained on massive high quality MOCAP data, when conditioned on a few key-frames, are capable of producing reasonably high quality and relatively long animation sequences, and thus save time for animators. This data driven approach leverages existing high quality MOCAP footage and is able to re-purpose it towards creating new animation sequences, providing a simple yet powerful user interface. In this new type of AI assisted workflow, the deep learning model fulfils a few critical functions. First, the model compresses a MOCAP dataset compactly, encoding it in its weights by developing an internal latent space representation of motion. Second, it exposes comprehensive inputs, enabling a familiar yet flexible and powerful user interface by accepting key-frame conditioning from the animator. Finally, the deep neural network queries its latent space with the information provided by the user and outputs a highly believable motion sequence matching natural motion statistics that is ideally blended into user-provided key-frames. It is clear that deep learning will continue to expand its reach and strive for domination in the field of 3D computer animation. 

The extent and velocity of this process depends significantly on the progress in improving accuracy, robustness and flexibility of deep motion models, which is the focus of our current paper. We 
propose a configuration of the deep motion in-betweening interpolator that has better accuracy than state-of-the-art approaches and is more robust to the out-of-distribution changes due to the local nature of the deep neural network inputs and outputs. 
Additionally, we contribute to democratizing the animation authoring by making our model code publicly available
\footnote{\url{https://github.com/boreshkinai/delta-interpolator}}.
A sample animation of our approach in action is shown in Figure~\ref{fig:animation_demo}.

\subsection{Summary of Contributions}

\begin{enumerate}
    \item We propose a novel deep $\Delta$-interpolator motion synthesis architecture. We make our model code public. It is worth noting that the most recent SOTA works in this domain still have no publicly released implementations~\cite{harvey2020robust, duan2021singleshot}.
    \item Recent study advocates a global coordinate system for deep motion synthesis~\cite{duan2021singleshot}. We provide empirical evidence of the advantage of the local motion synthesis operation by achieving SOTA results on two public datasets (LaFAN1 and Anidance).
    \item We make our processed data public along with the data-loaders and evaluation scripts, unlike other works, lowering the entry barrier to this domain.
\end{enumerate}

\section{Related Work}

Conditional motion synthesis, and more specifically motion completion, is the process of filling the missing frames in a temporal sequence given a set of key-frames. Early work focused on generating missing in-between frames by integrating key-frame information into space-time  models~\cite{witkin1988spacetime, ngo1993spacetime}. The next generation of in-betweening works developed a more sophisticated probabilistic approach to the constrained motion synthesis, formulating it as a \emph{maximum a posteriori} (MAP) problem~\cite{chai2007constraint} or a Gaussian process~\cite{grochow2004style, wang2007gaussian}. Over the past decade, deep learning has become the \emph{modus operandi}, significantly outperforming probabilistic approaches. Due to the temporal nature of the task, RNNs have dominated the field compared to other neural network architectures~\cite{holden2016framework, harvey2018recurrent}. The latest RNN work by~\citet{harvey2020robust} focuses on augmenting a Long Short Term Memory (LSTM) based architecture with time-to-arrival embeddings and a scheduled target noise vector, allowing the system to be robust to target distortions. \citet{gelejein2021lightweight} propose a lightweight and simplified recurrent architecture that maintains good performance, while significantly reducing the computational complexity. Beyond RNN models, motion has been modelled by 1D temporal convolutions~\cite{holden2015learning}. More recent work has relied on Transformer based models predicting the entire motion in one pass unlike auto-regressive recurrent models~\citep{devlin2018bert,duan2021singleshot}. 

Motion in-betweening is an instance of conditional motion generation, in which the conditioning signal has the form of key-frames. Therefore, methods using other types of conditioning to generate motion are related to ours. These range from audio conditioned human motion generation such as creating dances for a musical piece~\cite{kao2020temporally, li2021learn}, to semantically conditioned motion~\cite{chuan2020action2motion, petrovich2021actionconditioned}.

While being similar to our setting, motion prediction works by~\cite{martinez2017rnn, wang2022velocity} are not directly applicable. First, they typically rely on dense key-frame conditioning prior to the beginning of the missing frames~\citep{wang2022velocity} and do not support conditioning on a future frame.
Additionally, to reduce the error they sometimes rely on costly Inverse Kinematics (IK) projections to preserve limb lengths~\cite{wang2022velocity} or allow limb dislocations~\cite{kaufmann2020}. In a high quality animation authoring setting the computational cost of IK or the error that can result from skipping IK are unsustainable. Furthermore, in classical animation workflows we have a sparse key-frame distribution across a sequence with gaps of regular intervals rather than contiguous blocks of key-frames followed by a large gap as in motion prediction.


\section{Problem Statement and Background} \label{sec:problem_statement}

For a skeleton with $J$ joints, we define the tensor $\vec{x}_t \in \Re^{J \times 9}$ fully specifying the pose at time $t$. It contains the concatenation of the global positions $\vec{p}_t \in \Re^{J \times 3}$ as well as rotations $\vec{r}_t \in \Re^{J \times 6}$. By default, the rotation of the root joint is defined in the global coordinate system, whereas the rest of the joints specify rotations relative to their parent joints, unless explicitly stated otherwise. 

For positions, we use a standard 3D Euclidean representation. For joint rotations, we use the robust ortho6D representation proposed by~\citet{zhou2019onthecontinuity}. This addresses the continuity issues of quaternion and Euler representations. 

Suppose a unit vector is defined as $\overrightarrow{\vec{u}} \equiv \vec{u} / \| \vec{u} \|_2$ and the vector cross product is $\vec{u} \times \vec{v} = \|\vec{u}\| \|\vec{v}\| \sin(\gamma) \overrightarrow{\vec{n}}$ ($\gamma$ is the angle between $\vec{u}$ and $\vec{v}$ in the plane containing them and $\overrightarrow{\vec{n}}$ is the normal to the plane). Given the ortho6D rotation $\vec{r}_{t,j} \in \Re^6$, the rotation matrix of joint $j$ at time $t$, $\vec{R}_{t,j}$, is: 
\begin{align} 
\begin{split} \nonumber
    \vec{R}_{t,j}^x &= \overrightarrow{\vec{r}_{t,j}[:3]}, \quad\quad
    \vec{R}_{t,j}^z = \overrightarrow{\vec{R}_{t,j}^x \times \vec{r}_{t,j}[4:]}, \\
    \vec{R}_{t,j}^y &= \vec{R}_{t,j}^z \times \vec{R}_{t,j}^x, \quad
    \vec{R}_{t,j} = \left[\vec{R}_{t,j}^x \   \vec{R}_{t,j}^y \  \vec{R}_{t,j}^z \right].
\end{split}
\end{align}


The model observes a subset of samples in the time window $\mathcal{T} = \{1, \ldots, T\}$ of total length $T$, defined as $\vec{X} \equiv \{ \vec{x}_t: t \in \mathcal{T}_{in} \}, \ \mathcal{T}_{in} \subset \mathcal{T}$. The model solves the in-betweening task on the subset $\mathcal{T}_{out} = \mathcal{T} \setminus \mathcal{T}_{in}$ of missing samples $\vec{Y} \equiv \{ \vec{x}_t: t \in \mathcal{T}_{out} \}$ by predicting global positions and rotations of the root joint and local rotations of the other joints. This is sufficient to fully specify the pose by computing global positions and rotations of all joints after the Forward Kinematics (FK) pass. The forward kinematics pass operates on a skeleton defined by the offset vectors $\vec{o}_j = [o_{x,j}, o_{y,j}, o_{z,j}]^\intercal$ specifying the displacement of each joint with respect to its parent joint when joint $j$ rotation is zero (the norm of the offset vector is the bone length, which we consider to be constant). Provided with the local offset vectors and rotation matrices of all joints, the global rigid transform of any joint $j$ is computed following the tree recursion from its parent joint $p(j)$:
\begin{align} \nonumber
\vec{G}_{t,j} = \vec{G}_{t,p(j)} \begin{bmatrix}
\vec{R}_{t,j} & \vec{o}_j \\
\vec{0} & 1
\end{bmatrix}.
\end{align}
The global transform matrix $\vec{G}_{t,j}$ of joint $j$ contains its global rotation matrix, $\vec{G}_{t,j}[1:3, 1:3]$, and its 3D global position, $\vec{G}_{t,j}[1:3, 4]$.

In this paper we consider the variation of the in-betweening problem defined in the LaFAN1 benchmark~\cite{harvey2018recurrent}: A neural network is provided with 10 past poses and one future pose, the task being to fill the frames in between.
We also consider the in-filling version of the problem as posed in Figure 4 of~\citet{duan2021singleshot_arxiv} using the Anidance benchmark, where key-frames are distributed throughout the sequence in regular intervals of one key-frame every 6 frames.

\section{Method} \label{sec:method}

\begin{figure}[!ht]
    \centering
    \includegraphics[width=\linewidth]{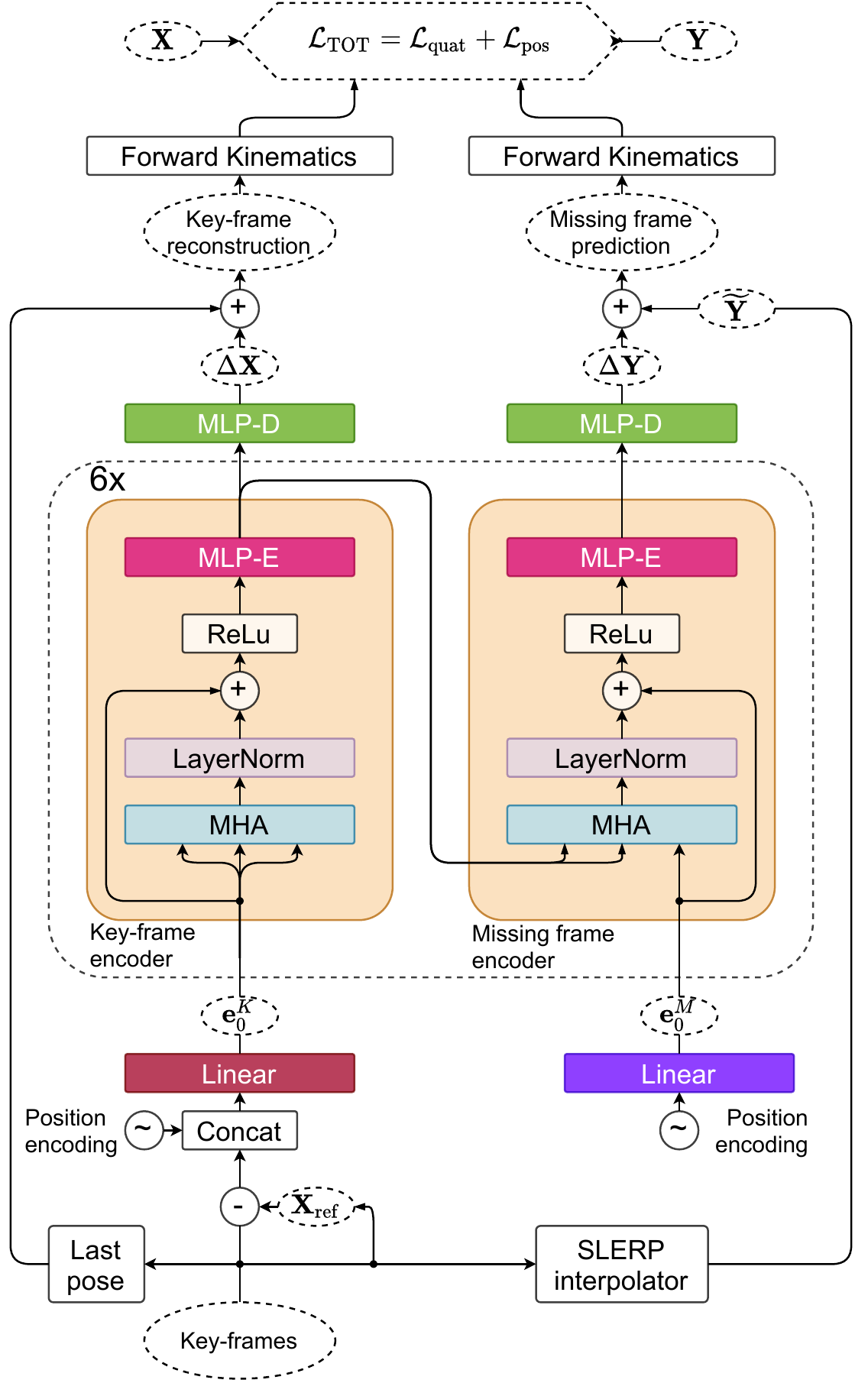}
    \caption{Proposed $\Delta$-interpolator architecture. Full-pose information from key-frames is referenced with respect to the root reference of the last frame in the past context, $\vec{X}_{\textrm{ref}}$, concatenated with positional encoding and linearly projected into model width $d$. Self-attention block encodes key-frame representations. Cross-attention block combines key-frame encodings with zero-initialized and positionally encoded templates of missing frames. Six encoder blocks are followed by a decoder MLP that produces $\Delta$-predictions of root position and joint rotations. Final prediction combines SLERP interpolator output with the $\Delta$-predictions of the neural network. 
    }
    \label{fig:delta_interpolator_architecture}
    \vspace{-1em}
\end{figure}

In this section we present the proposed $\Delta$-interpolator approach and its ingredients. We first introduce the spherical linear interpolator (SLERP) serving as the foundation for the proposed delta principle. We further discuss the details of the proposed neural network architecture, including the composition of its inputs and outputs as well as the mathematical details of the neural architecture. A high level block diagram is depicted in Fig.~\ref{fig:delta_interpolator_architecture}. We close the section with the detailed analysis of the architectural novelty of the proposed method with respect to the most closely related work by~\citet{duan2021singleshot} and the description of loss terms.

\subsection{The SLERP Interpolator Baseline}

Common tool-chains in the animation industry often offer the linear interpolator and blending curves as a tool for motion in-betweening. The curves require a significant amount of manual tweaking to make high quality motion transitions. The linear interpolator generates inbetween root motion via linear interpolation and inbetween local rotations via spherical linear interpolation (SLERP) in the quaternion space~\cite{shoemake1985animating}. We denote $\widetilde{\vec{Y}}$ the SLERP interpolator output computed based on $\vec{X}$. 

\subsection{The $\Delta$-Interpolator}

We propose the $\Delta$-interpolator approach in which the output of the neural network $f_{\theta}$ with parameters $\theta$ is defined as a delta w.r.t. the prediction of the linear interpolator, and its input is defined as a delta w.r.t. a reference $\vec{X}_{\textrm{ref}} \in \Re^{9}$. The latter can be defined, for example, as the concatenation of the 3D coordinate and 6D rotation of the root joint of the last history frame. The $\Delta$-solution, $\Delta\vec{Y}$:
\begin{align} 
\Delta\vec{Y} &\equiv f_{\theta}(\vec{X} - \vec{X}_{\textrm{ref}}), \label{eqn:delta_interpolator_in} \\
\widehat{\vec{Y}} &= \widetilde{\vec{Y}} + \Delta\vec{Y} \label{eqn:delta_interpolator_out},
\end{align}
belongs to the space local to the reference frame implicitly defined by the prediction of the linear interpolator, $\widetilde{\vec{Y}}$. Additionally, the input to $f_{\theta}$ is localized with respect to the reference frame $\vec{X}_{\textrm{ref}}$ defined based on the information available in $\vec{X}$. The intuition behind the proposed $\Delta$-interpolator is two-fold. First, if the linear interpolator provides a good initial prediction $\widetilde{\vec{Y}}$, solving the $\Delta$-interpolation task should be easier, requiring less training data and resulting in better generalization accuracy. Second, since the output is defined in the reference frame relative to the linear interpolator, it provides the opportunity for defining the input relative to a reference $\vec{X}_{\textrm{ref}}$ without loss of information.\footnote{Note that an approach that predicts absolute global positions would necessarily incur information loss if its input was made local.} As a result, the $\Delta$-interpolator neural network does not have to deal with any absolute domain shifts (either at the input or at the output). This improves its ability to cope with any global reference frame shifts and out-of-distribution domain shifts arising between training and inference scenarios. As an example, suppose the test data is defined in a reference frame that is shifted with respect to the original training reference frame by an arbitrary constant delta $\Delta \vec{p}$. In our experience, even relatively small $\Delta \vec{p}$, comparable to the span of the skeleton, is enough to significantly degrade performance. In contrast, it is easy to show that the $\Delta$-interpolator defined in equations (\ref{eqn:delta_interpolator_in}-\ref{eqn:delta_interpolator_out}) is completely insensitive to it. First, the domain shift in $\vec{X}$ will be compensated by that in $\vec{X}_{\textrm{ref}}$ and the input to the neural network will essentially be the same with or without $\Delta \vec{p}$. Moreover, the domain shift will be added  to the output of the $\Delta$-interpolator in equation~(\ref{eqn:delta_interpolator_out}) via $\widetilde{\vec{Y}}$, resulting in the correct reference shift presented at the output.

\subsection{The Neural Network}
In this section we describe the details of the proposed neural architecture, graphically depicted in Fig.~\ref{fig:delta_interpolator_architecture}. We start by defining some standard notions from the Transformer literature. We use the Transformer approach with a multi-head self-attention encoder for the key-frames and a multi-head cross-attention encoder for the missing frames. Following the original Transformer formulation~\cite{vaswani2017attention}, the attention unit is defined as:
\begin{align} \label{eqn:attention}
    \att(\vec{Q}, \vec{K}, \vec{\vec{V}}) = \softmax(\vec{Q} \vec{K}^T / \sqrt{d}) \vec{V},
\end{align}
where $d$ is model-wide hidden dimension width. The multi-head attention unit is defined as:
\begin{align}
    \vec{h}_i &= \att(\vec{Q} \vec{W}^{Q}_{i}, \vec{K} \vec{W}^{K}_{i}, \vec{\vec{V}} \vec{W}^{V}_{i}), \nonumber \\
    \mha(\vec{Q}, \vec{K}, \vec{\vec{V}}) &= \concat(\vec{h}_1, \ldots, \vec{h}_h) \vec{W}^{O}. \nonumber
\end{align}
Next, we describe the details of the proposed architecture.

\subsubsection{Inputs} 
As mentioned previously, the input of the network is localized w.r.t. $\vec{X}_{\textrm{ref}}$. We define $\vec{X}_{\textrm{ref}}  \in \mathbb{R}^{|\mathcal{T}_{in}| \times J \times 9}$ by (i) setting all its position components along time and joint axes to the root position at the last available frame; and (ii) setting all its rotation components along time and joint axes to the root rotation at the last available frame. Furthermore, $f_{\theta}$ accepts (i) $\vec{e}_{0}^{\textrm{K}}$: the key-frame embedding initialized as the concatenation of $\vec{X} - \vec{X}_{\textrm{ref}}$ with the learnable input position encoding, linearly projected into model width $d$; as well as (ii) $\vec{e}_{0}^{\textrm{M}}$: the missing frame template initialized as the learnable output position encoding, linearly projected into model width $d$. Input and output position embeddings are shared and consist of a learnable vector of size $32$ for each possible frame index $t$. 

\subsubsection{Key-frame encoder} 
The key-frame encoder is described by the following recursion over $\ell \in (0, R]$ encoder residual blocks taking $\vec{e}_{0}^{\textrm{K}}$ as input:
\begin{align}
    \vec{e}_{\ell}^{\textrm{K}} &= \mha_{\ell}(\vec{e}_{\ell-1}^{\textrm{K}}, \vec{e}_{\ell-1}^{\textrm{K}}, \vec{e}_{\ell-1}^{\textrm{K}}), \nonumber \\
    \vec{e}_{\ell}^{\textrm{K}} &= \relu(\layernorm_{\ell}(\vec{e}_{\ell}^{\textrm{K}} + \vec{e}_{\ell-1}^{\textrm{K}})), \nonumber \\
    \vec{e}_{\ell}^{\textrm{K}} &= \mlp_{\ell}(\vec{e}_{\ell}^{\textrm{K}}). \nonumber 
\end{align}
The purpose of the key-frame encoder is to create a deep representation of key-frames that can be further cross-correlated with the templates of missing frames in the missing frame encoder. Using a separate encoder for key frames is computationally efficient. Typically, the number of key frames is low and they encapsulate all the information to infer the missing in-between frames. Therefore, processing them in the same self-attention block as the missing frames, as proposed by~\citet{duan2021singleshot}, is unnecessary from the information processing standpoint and cost-ineffective from the computational standpoint. See Section~\ref{ssec:architectural_novelty} for an extended discussion.

\subsubsection{Missing frame encoder} 
The missing frame encoder iterates over $R$ residual blocks, cross-processing the key- and missing frame representations. At each level $\ell$, it accepts the key-frame encodings $\vec{e}_{\ell}^{\textrm{K}}$ created by the key-frame encoder and the missing frame encodings from the previous level, $\vec{e}_{\ell-1}^{\textrm{M}}$. The first level $\ell=1$ accepts the missing frame template, $\vec{e}_{0}^{\textrm{M}}$.
\begin{align}
    \vec{e}_{\ell}^{\textrm{M}}  &= \mha_{\ell}(\vec{e}_{\ell-1}^{\textrm{M}}, \vec{e}_{\ell}^{\textrm{K}}, \vec{e}_{\ell}^{\textrm{K}}), \nonumber \\
    \vec{e}_{\ell}^{\textrm{M}} &= \relu(\layernorm_{\ell}(\vec{e}_{\ell}^{\textrm{M}} + \vec{e}_{\ell-1}^{\textrm{M}})), \nonumber \\
    \vec{e}_{\ell}^{\textrm{M}} &= \mlp_{\ell}(\vec{e}_{\ell}^{\textrm{M}}). \nonumber 
\end{align}
Obviously, the missing frame encoder is  where the known key-frames meet the unknown missing in-between frames. Note that the common self-attention block in~\citep{duan2021singleshot} accepts both key-frames and missing frames, cross-polluting their representations. Our architecture implements a different information processing flow, emphasizing the cause and effect relationship between inputs and outputs in the in-betweening problem. Hence, in the case of our architecture design, the information flows in a directed fashion from the key-frames that are provided by the user as a conditioning signal towards the missing frames that are inferred at the output.

\begin{table*}[!ht]
\centering
\caption{Key Empirical Results on LaFAN1 dataset. Lower score is better.}
\begin{tabularx}{\textwidth}{l@{\extracolsep{\fill}} ccccccccc}
&\multicolumn{3}{c}{\textbf{L2Q}} &\multicolumn{3}{c}{\textbf{L2P}} &\multicolumn{3}{c}{\textbf{NPSS}} \\
\cmidrule(lr){2-4}\cmidrule(lr){5-7}\cmidrule(lr){8-10}
Length & 5 & 15 & 30 & 5 & 15 & 30 & 5 & 15 & 30\\
\cmidrule(lr){1-10}
Zero Velocity & 0.56 & 1.10 & 1.51 & 1.52 & 3.69 & 6.60 & 0.0053 & 0.0522 & 0.2318 \\
SLERP Interpolation & 0.22 & 0.62 & 0.98 & 0.37 & 1.25 & 2.32 & 0.0023 & 0.0391 & 0.2013 \\
\cmidrule(lr){1-10}
Autoencoder \cite{kaufmann2020} & 0.49 & 0.60 & 0.78 & 0.84 & 1.07 & 1.53 & 0.0048 & 0.0345 & 0.1454 \\
HHM-VAE \cite{li2021taskgeneric} & 0.24 & 0.54 & 0.94 & N/A & N/A & N/A & N/A & N/A & N/A \\ 
$\textrm{TG}_{rec}$ (rec. loss) \cite{harvey2020robust} & 0.21 & 0.48 & 0.83 & 0.32 & 0.85 & 1.82 & 0.0025 & 0.0304 & 0.1608 \\
$\textrm{TG}_{complete}$ (rec. \& adv. loss) \cite{harvey2020robust} & 0.17 & 0.42 & 0.69 & 0.23 & 0.65 & 1.28 & 0.0020 & 0.0258 & 0.1328 \\
SSMCT (local) \cite{duan2021singleshot}  & 0.17 & 0.44 & 0.71 & 0.23 & 0.74 & 1.37 & 0.0019 & 0.0291 & 0.1430\\
SSMCT (global) \cite{duan2021singleshot}  & 0.14 & 0.36 & 0.61 & 0.22 & 0.56 & 1.10 & 0.0016 & 0.0234 & 0.1222 \\
$\Delta$-Interpolator (Ours)  & \bf{0.11}	& \bf{0.32} &	\bf{0.57} &	\bf{0.13} &	\bf{0.47} &	\bf{1.00} &	\bf{0.0014} &	\bf{0.0217} &	\bf{0.1217} \\
\bottomrule
\end{tabularx}%
\label{table:key_results}
\end{table*}%


\subsubsection{Decoder} 
The decoder is an MLP that maps the representations of key-frames and missing frames, $\vec{e}_{L}^{\textrm{K}}$ and $\vec{e}_{L}^{\textrm{M}}$, into the predictions of full poses of all frames in the sequence:
\begin{align}
    \Delta\vec{Y}, \Delta\vec{X} = \mlp_{\textrm{D}}(\vec{e}_{L}^{\textrm{M}}), \mlp_{\textrm{D}}(\vec{e}_{L}^{\textrm{K}}). \nonumber
\end{align}

Note that the MHA and MLP blocks are shared between the key-frame and missing frame encoder blocks; similarly, the same decoder is used to output key-frame reconstructions and missing frame predictions.

\subsubsection{Outputs} 
The $\Delta$-predictions $\Delta\vec{Y}, \Delta\vec{X}$ supplied by the decoder contain a $J \times 6$ rotation prediction (global rotation of the root plus local rotations of all other joints in their parent-relative coordinate systems) and a $1 \times 3$ global root joint position prediction. These are further subjected to the skeleton forward kinematics pass to derive the full pose missing frame predictions $\widehat{\vec{Y}} \in \mathbb{R}^{|\mathcal{T}_{out}| \times J \times 9}$ and the key-frame reconstructions $\widehat{\vec{X}} \in \mathbb{R}^{|\mathcal{T}_{in}| \times J \times 9}$ containing  $J \times 6$ global rotations and $J\times 3$ global positions of all joints.
\begin{align}
\widehat{\vec{Y}} &= \fk(\widetilde{\vec{Y}} + \Delta\vec{Y}); \quad \widehat{\vec{X}} = \fk(\widetilde{\vec{X}} + \Delta\vec{X}). \nonumber
\end{align}

\subsection{Architectural novelty with respect to~\cite{duan2021singleshot}.} \label{ssec:architectural_novelty}

The architecture in~\cite{duan2021singleshot} is based on a single self-attention module; key-frames concatenated with the linear interpolator predictions are used as input to the self-attention; the key-frames and the missing frames are marked with the key-frame encoding so that the neural network knows which frames to predict. We experimentally confirm in Section~\ref{sec:experimental_results} several important observations. First, when we use self-attention for the key-frames and cross-attention for the missing frames, the key-frame encoding is not needed. Second, we show that MHA and MLP blocks can be shared in the self-attention and cross-attention blocks. Weight sharing is important for reducing the model size. Third, the architecture of~\citet{duan2021singleshot} is not operational without the interpolator providing input initialization. Therefore, using this architecture in contexts such as motion prediction is not viable. In our case, neither $\vec{X}_{\textrm{ref}}$ nor $\widetilde{\vec{Y}}$ has to be based on a linear interpolator. We show that our approach can use the last frame reference both for $\vec{X}_{\textrm{ref}}$ and $\widetilde{\vec{Y}}$, while the missing frame input is initialized with zeros. This makes our approach significantly more general (\emph{e.g.}, also suitable for motion prediction applications). Finally, processing key-frames via self-attention and missing frames via cross-attention is more computationally efficient. Indeed, for a problem with $n_{\textrm{k}}$ key-frames and $n_{\textrm{in}}$ inbetween frames, attention complexity (\emph{i.e.}, the complexity of equation~\eqref{eqn:attention}) of~\cite{duan2021singleshot} scales as $(n_{\textrm{k}} + n_{\textrm{in}})^2$, whereas our approach's attention complexity scales as $n_{\textrm{k}}^2 + n_{\textrm{k}} n_{\textrm{in}}$. The ratio of the two quantities is $1 + n_{\textrm{in}} / n_{\textrm{k}}$. Suppose $n_{\textrm{in}} = 30$ and $n_{\textrm{k}} = 11$, we have that the attention complexity of our approach is asymptotically $1 + 30/11 = 3.7$ times smaller.

\subsection{Losses}
The neural network is supervised using a combination of quaternion and position losses, similar to~\cite{duan2021singleshot}:
\begin{align}
\mathcal{L}_{\textrm{TOT}} = \mathcal{L}_{\textrm{quat}} + \mathcal{L}_{\textrm{pos}}. \nonumber
\end{align}
The position loss is L1 loss defined both on missing and key-frames. In the case of missing frames we call it predictive position loss. In the case of key-frames we call it the reconstruction position loss, as then the network acts as an auto encoder. $\vec{X}_{\textrm{pos}}$ and $\vec{Y}_{\textrm{pos}}$ are position only components of $\vec{X}$ and $\vec{Y}$, respectively (\emph{e.g.}, $\vec{X}_{\textrm{pos}} \equiv \{ \vec{p}_t: t \in \mathcal{T}_{in} \}$).
\begin{align} \label{eqn:position_loss}
\mathcal{L}_{\textrm{pos}} = \| \vec{Y}_{\textrm{pos}} - \widehat{\vec{Y}}_{\textrm{pos}} \|_1 + \| \vec{X}_{\textrm{pos}} - \widehat{\vec{X}}_{\textrm{pos}} \|_1. 
\end{align}

Rotational output is supervised with the L1 loss based on quaternion representations $\widehat{\vec{X}}_{\textrm{quat}} \in \mathbb{R}^{B\times|\mathcal{T}_{in}|\times 4}$ and $\widehat{\vec{Y}}_{\textrm{quat}} \in \mathbb{R}^{B\times|\mathcal{T}_{out}|\times 4}$ derived from the ortho6D predictions of global rotations contained in $\widehat{\vec{X}}$ and $\widehat{\vec{Y}}$:
\begin{align} \label{eqn:quaternion_loss}
\mathcal{L}_{\textrm{quat}} = \| \vec{Y}_{\textrm{quat}} - \widehat{\vec{Y}}_{\textrm{quat}} \|_1 + \| \vec{X}_{\textrm{quat}} - \widehat{\vec{X}}_{\textrm{quat}} \|_1. 
\end{align}
The L1 norm here is taken over the last tensor dimension whereas the two leading dimensions are average pooled.

\begin{table}[!htb]
\centering
\caption{Results on Anidance dataset. Lower score is better.}
\begin{tabularx}{\columnwidth}{l@{\extracolsep{\fill}} ccc}
&\multicolumn{3}{c}{\textbf{L2P}} \\
\cmidrule(lr){2-4}
Length & 5 & 15 & 30 \\
\cmidrule(lr){1-4}
Zero Velocity & 2.44 & 5.15 & 6.89 \\
LERP Interpolation & 0.94 & 3.06 & 4.84 \\
\cmidrule(lr){1-4}
Autoencoder \cite{kaufmann2020} & 3.57 & 3.69 & 3.93 \\ 
SSMCT \cite{duan2021singleshot}  & 0.84 & 1.46 & 1.64 \\
$\Delta$-Interpolator (w/o $\Delta$-mode)  & 0.93	& 1.02 &	1.13\\
$\Delta$-Interpolator (ours)  & \bf{0.60}	& \bf{0.74} &	\bf{1.01}\\

\bottomrule
\end{tabularx}%
\label{table:anidance_results}
\end{table}%

\section{Experimental Results} \label{sec:experimental_results}

In this section we present our empirical results. The section starts with the description of the datasets, as well as the benchmark and metrics used for the qualitative evaluation of our model. We then carefully describe the details of training and evaluation setups as well as the computational budget used to produce our results. Our key results are presented in Tables~\ref{table:key_results} and~\ref{table:anidance_results}, demonstrating the state-of-the-art performance of our method relative to existing work. Finally, we conclude this section by describing the ablation studies we conducted. The results of ablations confirm the significance and flexibility of our $\Delta$-interpolator approach.

\subsection{Dataset and Benchmarking} 
We empirically evaluate our method on the well-established motion synthesis benchmarks LaFAN1~\cite{harvey2020robust} and Anidance~\cite{tang2018}. LaFAN1 consists of 496,672 motion frames captured in a production-grade MOCAP studio. The footage is captured at 30 frames per second for a total of over 4.5 hours of content. There are 77 sequences in total with 15 action categories that include a diverse set of activities such as walking, dancing, and fighting. Each sequence contains actions performed by one of 5 subjects (MOCAP actors). Sequences with subject 5 are used as the test set. We build the training and test sets following~\citet{harvey2020robust}, sampling smaller sliding windows from the original long sequences. We sample our training windows from sequences acted by subjects 1-4 and retrieve windows of 50 frames with an offset of 20 frames. We repeat a similar exercise for the test set, sampling sequences of subject 5 at every 40 frames, yielding 2232 test windows of 65 frames. We also extract the same statistics used for data normalization as outlined in~\citep{harvey2020robust}, applying the same data normalization procedures. 
The Anidance dataset was originally proposed as a music-to-dance generation dataset~\citep{tang2018}. Anidance contains 61 dance sequences from 4 genres shot at 25 frames per second. In total, the dataset is composed of 101,390 frames of global positional coordinates of the skeletal joints (no joint rotations are provided). \citet{duan2021singleshot} discarded the audio features and used the dataset as a motion completion benchmark following pre-processing steps similar to LaFAN1 with windows and offset of size 128 and 64 frames respectively for both train and test splits. 

We report our key empirical results in Tables~\ref{table:key_results},~\ref{table:anidance_results}, demonstrating state-of-the-art performance of our method. Our results are based on 8 different random seed runs of the algorithm and metric values averaged over the 10 last optimization epochs. Following previous work, we consider L2Q (the global quaternion squared loss~\eqref{eqn:l2q}), L2P (the global position squared loss~\eqref{eqn:l2p}) and NPSS (the normalized power spectrum similarity score)~\cite{gopalakrishnan2019neural}. Table~\ref{table:anidance_results} does not contain L2Q and NPSS metrics because the Anidance dataset does not contain angular information.
\begin{equation} \label{eqn:l2q}
    \text{L2Q} = \frac{1}{|\mathcal{D}|}\frac{1}{|\mathcal{T}_{out}|} \sum_{s \in \mathcal{D}} \sum_{t \in \mathcal{T}_{out}}  ||\widehat{\vec{q}}_t^s - \vec{q}_t^s ||_2, 
\end{equation}
\begin{equation} \label{eqn:l2p}
    \text{L2P} = \frac{1}{|\mathcal{D}|}\frac{1}{|\mathcal{T}_{out}|} \sum_{s \in \mathcal{D}} \sum_{t \in \mathcal{T}_{out}}  ||\widehat{\vec{p}}_t^s - \vec{p}_t^s ||_2,
\end{equation}
where $\vec{q}_t\in\mathbb{R}^{J \times 4}$ represents the quaternion vector of all skeletal joints at time $t$, $\vec{p}_t\in\mathbb{R}^{J\times 3}$ denotes the global position vectors, superscript $s$ denotes sequences in the test dataset $\mathcal{D}$. We measure the metrics for scenarios with $|\mathcal{T}_{out}| \in \{5, 15, 30\}$ missing in-between frames.  

\begin{table*}[!ht]
\centering
\caption{Ablation of the $\Delta$-interpolation regime based on LaFAN1 dataset. Lower score is better.}
\centering
\begin{tabular}{llccccccccc}
\multicolumn{2}{c}{\textbf{$\Delta$-mode}} &\multicolumn{3}{c}{\textbf{L2Q}} &\multicolumn{3}{c}{\textbf{L2P}} &\multicolumn{3}{c}{\textbf{NPSS}} \\
\cmidrule(lr){3-5}\cmidrule(lr){6-8}\cmidrule(lr){9-11}
Input & Output & 5 & 15 & 30 & 5 & 15 & 30 & 5 & 15 & 30\\
\cmidrule(lr){1-11}
Last & I & \bf{0.11}	& \bf{0.32} &	\bf{0.57} &	\bf{0.13} &	\bf{0.47} &	\bf{1.00} &	\bf{0.0014} &	\bf{0.0217} &	\bf{0.1217} \\
Last & Last & 0.12 & 0.33 & 0.58 & 0.14 & 0.49 & 1.01 & 0.0015 & 0.0221 & \bf{0.1217} \\
No & No &  0.15 & 0.35 & 0.62 & 0.22 & 0.56 & 1.17 & 0.0017 & 0.0238 & 0.1300 \\
No & I &  \bf{0.11}  & \bf{0.32} & 0.59 & \bf{0.13} & 0.48 & 1.09 & \bf{0.0014} & 0.0221 & 0.1252 \\
No & Last &  0.12 & 0.33 & 0.59 & 0.14 & 0.51 & 1.12 & 0.0015 & 0.0227 & 0.1245 \\ 
\cmidrule(lr){1-11}
\end{tabular}
\label{tab:ablation_delta}
\end{table*}

\subsection{Training and Hyperparameters} 
\textbf{The training loop} is implemented in PyTorch~\cite{paszke2019pytorch} using the Adam optimizer~\cite{kingma2015adam}. The learning rate is warmed to 0.0002 during the first 50 epochs using a linear schedule and it steps down by a factor of 10 at epoch 250; training is continued until epoch 300. Most 3D geometry operations
are handled by pytorch3d~\citep{ravi2020pytorch3d}.

\textbf{Batch sampling.} Following previous work, we split the entire dataset into windows of maximum length $T_{\max}$ (see supplementary for the details of building the datasets). To construct each batch of size 64, we set the number of the start key-frames to be 10 for LaFAN1, 1 for Anidance and the number of the end key-frames to be 1 for both datasets. We then sample the number of in-between frames from the range [5, 39] without replacement. We employ weighted sampling, and the weight associated with the number of in-between frames $n_{\textrm{in}}$ is set to be inversely proportional to it, $w_{n_{\textrm{in}}} = 1 / n_{\textrm{in}}$. This weighting prevents overfitting on the windows with a large number of in-between frames.
Since Anidance has a periodic keyframe distribution, once the transition length has been sampled we provide every 6-th frame as an input frame along with the first and last frames of the sampled window and mask all other frames.

\textbf{Hyperparameters and compute requirements} are discussed in detail in the supplementary. The neural network encoder has 6 residual blocks $R$ of width 1024 with 8 MHA heads and 3 MLP layers each. The decoder's MLP has one hidden layer. Each row in our empirical tables is based on 8 different random seed runs of the algorithm and metric values averaged over 10 last epochs and 8 different random seeds. 

\subsection{Ablation Studies}

\subsubsection{$\Delta$-regime ablation}

The ablation of the $\Delta$-regime is presented in Table~\ref{tab:ablation_delta}, which compares our proposed approach ($\Delta$-Interpolator) against a few variants, applying different $\Delta$ configurations at the input and output of the neural network. We explored three $\Delta$ configurations and one configuration with no delta-regime at either input or output. Configuration ``I'' indicates that the output is $\Delta$ w.r.t. to the linear interpolator. Configuration ``L'' indicates for the input that it is in the $\Delta$-mode w.r.t $\vec{X}_{\textrm{ref}}$ and for the output that it is in the $\Delta$-mode w.r.t. the Zero-Velocity model (last history frame). Finally, configuration ``No'' indicates no $\Delta$-mode (\emph{i.e.}, at the output the neural network predicts \emph{global} root position directly and at the input $\vec{X}_{\textrm{ref}}$ is \emph{not} subtracted). 

Our proposed $\Delta$-Interpolator corresponds to the configuration (Last I) in the first row of Table~\ref{tab:ablation_delta}. The first alternative in the ablation study (Last Last) is different in that instead of relying on SLERP as a baseline, it uses the last known frame as a baseline, both at the input and at the output, making the neural network operation completely local w.r.t. the reference frame implied by the last known frame. We can see that this leads to a small but consistent deterioration of metrics compared to the proposed $\Delta$-interpolator. It is worth pointing out, however, that this variant does not at all rely on the SLERP interpolator. Therefore, it is viable to achieve very impressive results simply using $\Delta$ interpolation with respect to the last known frame. We noticed that the SLERP interpolator may be computationally quite demanding and hard to optimize on a GPU. Therefore, in applications where computation is a bottleneck, the (Last Last) configuration may be attractive. The third row of Table~\ref{tab:ablation_delta} is the configuration that does not rely on $\Delta$ interpolation at all. We can see that it has noticeably degraded performance. Still, compared to existing methods in Table~\ref{table:key_results} performance is very competitive, validating our neural network architecture design. Comparing the first and the third rows in Table~\ref{tab:ablation_delta} proves the effectiveness of the proposed $\Delta$-Interpolator approach. Rows 4-5 ablate the use of the $\Delta$-mode at the input of the deep neural network. We can see that (i) the input $\Delta$-mode helps when the output $\Delta$-mode is set to the last frame; (ii) the positive impact of the input $\Delta$-mode is more pronounced when the transition length is longer; and (iii) $\Delta$-mode at the input has positive effect both for the last frame (Last) and the Interpolator (I) configurations of the output $\Delta$-mode.

\begin{figure}[!t]
    \centering
    \animategraphics[loop,autoplay,controls,width=\linewidth]{24}{animation2/GT_INT_NT_DT_noPP_fallAndGetUp_frame1009_int30-}{0}{48}
    \caption{Robustness of the $\Delta$-interpolator (right, green), w.r.t. the out-of-distribution operation. Ground truth motion (left, white), SLERP interpolator (middle-left, yellow), proposed model with both input and output delta modes disabled (middle-right, blue). All models are fed with the data that are not subjected to the input normalization applied during training. Positional errors are indicated using red mask. $\Delta$-interpolator is not affected by distribution shift. \textbf{The animation is best viewed in Adobe Reader.}}
    \label{fig:delta-Interpolator_robustness}
    \vspace{-2em}
\end{figure}

Qualitatively, the robustness of the proposed $\Delta$-interpolator is demonstrated in Fig.~\ref{fig:delta-Interpolator_robustness}. Following the training protocol of~\citet{harvey2020robust}, the neural network is trained with the data normalized via XZ-center and Y-rotate operations. Normally, the same operations would have to be applied at inference time to ensure proper operation of all networks operating with global inputs and outputs, including~\citep{harvey2020robust, duan2021singleshot} and our approach with disabled $\Delta$-mode, otherwise their failure is unavoidable, which we can see clearly in Fig.~\ref{fig:delta-Interpolator_robustness}. In contrast, the $\Delta$-interpolator demonstrates robust out-of-distribution operation, insensitive to the distribution shift induced in this experiment.



\section{Discussion of Findings} \label{sec:discussion_of_findings}
Our results reveal SoTA performance of the $\Delta$-interpolator, defining both input and output of its deep neural network in the local coordinate frame derived from key-frames. This is opposite to the findings of~\citet{duan2021singleshot}, who advocate the use of the global reference frame for neural interpolation. Our approach produces more accurate results and makes the neural network robust w.r.t. out-of-distribution domain shifts due to its local nature. We believe this is an important contribution to the methodology of developing robust and accurate in-betweening algorithms. 

Our ablation studies demonstrate that our approach without $\Delta$-regime is significantly worse (although competitive against~\cite{duan2021singleshot}), confirming the importance of the proposed $\Delta$-interpolation approach. We additionally show that the last known frame can be used in lieu of SLERP as a reference, resulting in only slightly worse performance. This obviates the dependency on SLERP while doing neural in-betweening, further validates the general $\Delta$-interpolation idea laid out in equations~\eqref{eqn:delta_interpolator_in} and \eqref{eqn:delta_interpolator_out}, and makes our approach suitable for forward motion prediction applications. 

The ablation studies (see supplementary for ablation of our proposed reconstruction loss) imply that different input and output referencing methods have a noticeable effect, opening up a direction for future research. In our view, answering the questions ``What is a simple and more optimal $\vec{X}_{\textrm{ref}}$?''; and ``What is a simple and more optimal $\widetilde{\vec{Y}}$?'' has the potential to improve predictive accuracy without inflating computational costs. 
We believe this is pointing the community towards (i) looking for more optimal auxiliary reconstruction losses, (ii)  smarter sampling of inputs, or (iii) employing data augmentation in conjunction with reconstruction as potentially promising ways to further improve generalization results.

\section{Conclusions} \label{sec:conclusions}
This paper solves the prominent 3D animation task of motion completion. We propose the in-betweening algorithm in which a deep neural network acts in the $\Delta$-regime predicting the residual with respect to a linear SLERP interpolator. We empirically demonstrate that this mode of operation leads to more accurate neural in-betweening results on the publicly available LaFAN1 and Anidance benchmarks. Additionally, the $\Delta$-regime implies stronger robustness with respect to out-of-distribution shifts as both the input and the output of the network can be defined in the reference frame local to the SLERP interpolator baseline. Moreover, we show that the last known frame can be used in lieu of the SLERP interpolator, further simplifying the implementation of the algorithm at a small accuracy cost. 


\bibliography{main}

\clearpage
\appendix

\begin{figure*}[t!]
    \centering
    \includegraphics[width=2\columnwidth]{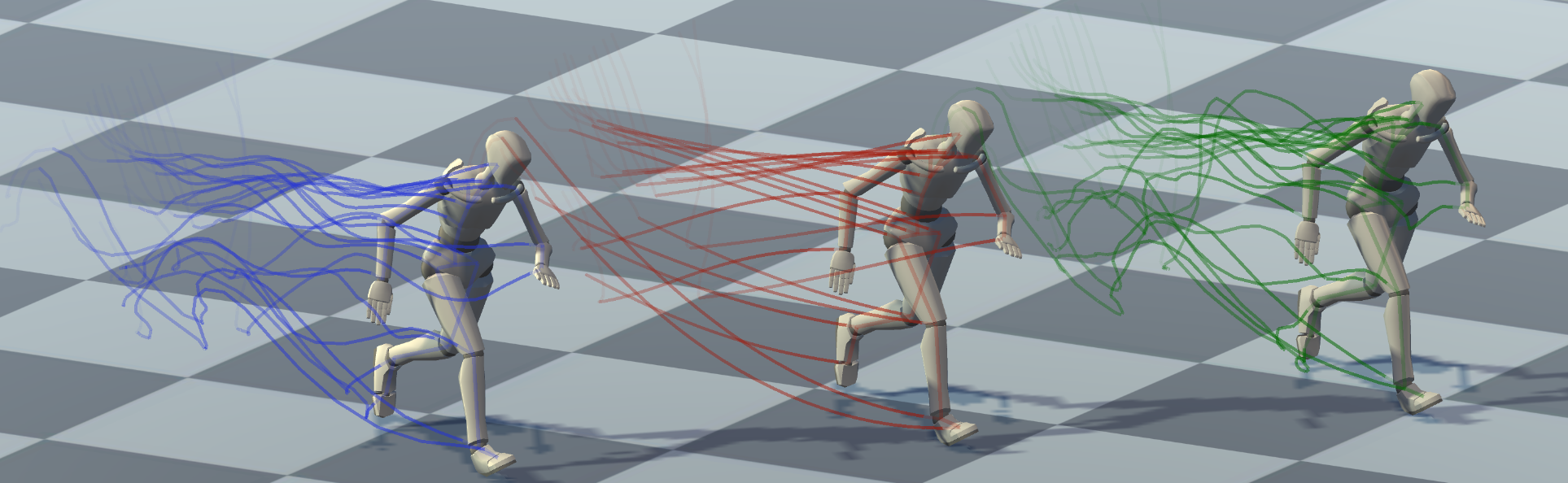}
    \caption{Comparison of human motion prediction. (Left): Our model. (Center): Linear interpolation. (Right): Ground Truth. Tracer lines represent evolution of joint locations over time.}
    \label{fig:bonus_pic}
\end{figure*}



\section{Ablation of Proposed Reconstruction Loss}
\label{ssec:ablation_of_reconstruction_loss}

Recall that the position and quaternion losses are:
\begin{align} \label{eqn:position_loss}
\mathcal{L}_{\textrm{pos}} = \| \vec{Y}_{\textrm{pos}} - \widehat{\vec{Y}}_{\textrm{pos}} \|_1 + \| \vec{X}_{\textrm{pos}} - \widehat{\vec{X}}_{\textrm{pos}} \|_1,
\end{align}
\begin{align} \label{eqn:quaternion_loss}
\mathcal{L}_{\textrm{quat}} = \| \vec{Y}_{\textrm{quat}} - \widehat{\vec{Y}}_{\textrm{quat}} \|_1 + \| \vec{X}_{\textrm{quat}} - \widehat{\vec{X}}_{\textrm{quat}} \|_1. 
\end{align}

The ablation study of the reconstruction loss terms is presented in Table~\ref{tab:ablation_loss_terms}. The second row in this table shows the generalization result with the reconstruction loss term removed. The reconstruction term includes two parts: $\| \vec{X}_{\textrm{pos}} - \widehat{\vec{X}}_{\textrm{pos}} \|_1$ from \eqref{eqn:position_loss} and $\| \vec{X}_{\textrm{quat}} - \widehat{\vec{X}}_{\textrm{quat}} \|_1$ from \eqref{eqn:quaternion_loss}. Interestingly, these terms do not directly penalize the errors on the missing frames, but rather only the reconstruction loss on the key frames. Also, the key frames are provided as inputs and it might be logical to think that the task of reconstructing them should be trivial. However, we can clearly see that penalizing the reconstruction error on the key frames provides positive regularizing effect showing in improved metrics measured on the test missing frames.

\begin{table*}[!ht]
\centering
\caption{Ablation of the loss terms based on LaFAN1 dataset. Lower score is better.}
\centering
\begin{tabular}{cccccccccc}
\multicolumn{1}{c}{} &\multicolumn{3}{c}{\textbf{L2Q}} &\multicolumn{3}{c}{\textbf{L2P}} &\multicolumn{3}{c}{\textbf{NPSS}} \\
\cmidrule(lr){2-4}\cmidrule(lr){5-7}\cmidrule(lr){8-10}
Reconstruction Loss & 5 & 15 & 30 & 5 & 15 & 30 & 5 & 15 & 30\\
\cmidrule(lr){1-10}
\checkmark & \bf{0.11}	& \bf{0.32} &	\bf{0.57} &	\bf{0.13} &	\bf{0.47} &	\bf{1.00} &	\bf{0.0014} &	\bf{0.0217} &	\bf{0.1217} \\
\ballot & 0.13 & 0.34 & 0.59 & 0.15 & 0.50 & 1.03 & 0.0015 & 0.0228 & 0.1247 \\
\cmidrule(lr){1-10}
\end{tabular}
\label{tab:ablation_loss_terms}
\end{table*}

\section{Hyperparameter Settings} \label{app:hyperparameter_settings}

\begin{table}[!ht]
    \centering
    \begin{tabular}{lc}
        \toprule
        Hyperparameter & Value  \\ 
        \midrule
        Epochs	& 300	\\ 
        Batch size & 64	\\
        Optimizer & Adam \\
        Learning rate, max & 2e-4  \\
        Warmup period, epochs & 50 \\
        Scheduler step size & 200 \\
        Scheduler gamma & 0.1 \\
        Dropout & 0.2 \\
        Losses & L1 \\
        Width ($d$) & 1024 \\
        MHA heads ($h$) & 8 \\
        Residual Blocks ($R$) & 6 \\
        Layers, encoder MLP ($L_{\textrm{E}}$) & 3 \\
        Layers, decoder MLP ($L_{\textrm{D}}$) & 2 \\
        Embedding dimensionality & 32 \\
        Augmentation & NONE \\
        \bottomrule
    \end{tabular}
    \vspace{1em}
    \caption{Hyperparameters of our model on both benchmarks.}
    \label{table:hyperparameter_settings}
\end{table}

The training loop is implemented in PyTorch~\cite{paszke2019pytorch} using Adam optimizer~\cite{kingma2015adam}. The learning rate is warmed to 0.0002 during the first 50 epochs using a linear schedule and it steps down by a factor of 10 at epoch 250; training is continued until epoch 300. Most 3D geometry operations (transformations across quaternions, ortho6d and rotation matrix representations, quaternion arithmetic) are handled by pytorch3d~\citep{ravi2020pytorch3d}. 

Following previous work, we split the entire dataset into windows of maximum length $T_{\max}$ (see Appendix~\ref{sec:dataset_details} for the details of building the dataset). To construct each batch (batch size is 64), we set the number of the start key-frames to be 10 and the number of the end key-frames to be 1. We then sample the number of in-between frames from the range [5, 39] without replacement. We employ weighted sampling, and the weight associated with the number of in-between frames $n_{\textrm{in}}$ is set to be inversely proportional to it, $w_{n_{\textrm{in}}} = 1 / n_{\textrm{in}}$. This weighting prevents overfitting on the windows with a large number of in-between frames. Shorter windows are sampled more often as they are more abundant and hence harder to overfit. Indeed, the number of unique non-overlapping sequences of a given total length $10 + 1 + n_{\textrm{in}}$ is approximately inversely proportional to $n_{\textrm{in}}$. Finally, given the total sampled sequence length $1+10+n_{\textrm{in}}$, the sequence start index is sampled uniformly at random in the range $[0, T_{\max}-(1+10+n_{\textrm{in}})]$.

The neural network encoder has 6 residual blocks $R$ of width 1024 with 8 MHA heads and 3 MLP layers each. The decoder's MLP has one hidden layer. All hyperparameter settings are provided in Table~\ref{table:hyperparameter_settings}. 

\section{Compute Requirements}\label{sec:compute}

Each row in our empirical tables is based on 8 different random seed runs of the algorithm and metric values averaged over 10 last optimization epochs and the 8 different random seed runs. One training run takes approximately 8 hours on a single NVIDIA M40 GPU on Dell PowerEdge C4130 server equipped  with 2 x Intel Xeon E5-2660v3 CPUs. It takes roughly 4 days on the aforementioned server to reproduce the LaFAN1 results presented in the main paper (6 rows x 8 runs x 8 hours / 4 GPUs).


\section{LaFAN1 Data Preparation} \label{sec:dataset_details}

Since the authors of~\cite{harvey2020robust} did not release a PyTorch compatible data loader we implement it on our own in accordance with their description of the sliding window technique and data normalization. To ensure that our approach is correct we implement the same basic baselines as the authors, also in PyTorch, including a zero-velocity model and spherical linear interpolator (SLERP). We then validate our implementation by obtaining the same values for the metrics on the zero-velocity and SLERP baselines as the ones reported in~\cite{harvey2020robust}. Note that our results in the main paper report zero-velocity and SLERP metrics based on our own implementation of the data loader and the algorithms.

As previously noted, the LaFAN1 benchmark\footnote{Accessible at: \url{https://github.com/ubisoft/ubisoft-laforge-animation-dataset}}~\cite{harvey2020robust} was provided to us in raw BVH format.
To make the data usable in the PyTorch and Unity pipeline we implemented, it was necessary to first convert it to a set of regular CSV files (one for each BVH animation) and to assign the animations to a Unity engine avatar for integrating our model in the engine as shown in Figure 1 of the main paper.
This section outlines the steps taken for this conversion:
\begin{enumerate}
    \item Download the original LAFAN1 source and follow the steps outlined in the repository (see footnote 5) to extract 77 BVH files and verify data integrity.
    \item To import the animations to the Unity engine we first need to convert them from BVH to FBX files. This is done in an open source, free computer graphics software called Blender\footnote{Available for download at: \url{https://www.blender.org/}}.
    \item Create a new Unity project and import all the animations as assets.
    \item We disable compression, unit conversion, keyframe reduction and enable baking of axis conversion. 
    These settings were chosen to maximize the animation quality.
    \item To make the animations viewable in Unity we select the clip group and add an avatar to the FBX animation clips.
    \item We create a dataset from the clip group with the following parameters set: Bone positions (world space), Bone rotations (local space), root reference (local space) and timestamps. We do not enable root motion.
    \item Finally, we use a proprietary animation baker component to the clip, link it to our skeleton characterization and select 30 FPS output. The baker provides CSV files for each of the 77 animations with time stamps.
\end{enumerate}

While our process results in high fidelity copies of the BHV animations in CSV format there are some small sources of error.
Specifically, steps 2 and 3 introduce most of the error since Blender has a right handed coordinate system whereas the Unity engine operates with left handed coordinates. The different handedness of the coordinate systems introduces a flip of the sign along the Z-axis (which we deal with by exporting negative Z-axis coordinates from Blender hence introducing another axis flip thus restoring the original coordinate orientation). However, this also changes the exported joint angular rotations which are challenging to fix and are hence left as is besides applying a quaternion discontinuity fix similar to~\cite{harvey2020robust}.

To validate that our approach produces high fidelity conversions from BVH to CSV we re-implement and compare the zero velocity and linear interpolation baselines and evaluate them on the same tasks as~\cite{harvey2020robust}. 
Since these dummy models have zero parameters if our pre-processing does not compromise data integrity we should obtain very similar values to what was reported in~\cite{harvey2020robust}.
As we can see in Table~\ref{tab:zeroParamComparison} the benchmark models perform within $\sim$ 1\% thus validating our pre-processing pipeline.

\begin{table*}[!ht]
\centering
\caption{Comparison of zero parameter models to verify data integrity of LaFAN1 benchmark.}
\begin{tabularx}{\textwidth}{l@{\extracolsep{\fill}} ccccccccc}
&\multicolumn{3}{c}{\textbf{L2Q}} &\multicolumn{3}{c}{\textbf{L2P}} &\multicolumn{3}{c}{\textbf{NPSS}} \\
\cmidrule(lr){2-4}\cmidrule(lr){5-7}\cmidrule(lr){8-10}
Length & 5 & 15 & 30 & 5 & 15 & 30 & 5 & 15 & 30\\
\cmidrule(lr){1-10}
Zero Velocity~\cite{harvey2020robust} & 0.56 & 1.10 & 1.51 & 1.52 & 3.69 & 6.60 & 0.0053 & 0.0522 & 0.2318 \\
Zero Velocity (ours) & 0.56 & 1.10 & 1.51 & 1.51 & 3.67 & 6.56 & 0.0053 & 0.0521 & 0.2324 \\
\cmidrule(lr){1-10}
SLERP Interpolation~\cite{harvey2020robust} & 0.22 & 0.62 & 0.98 & 0.37 & 1.25 & 2.32 & 0.0023 & 0.0391 & 0.2013 \\
SLERP Interpolation (ours) & 0.22 & 0.62 & 0.97 & 0.37 & 1.24 & 2.28 & 0.0023 & 0.0390 & 0.2061 \\
\end{tabularx}%
\label{tab:zeroParamComparison}
\end{table*}%

\section{Anidance Data Preparation} \label{sec:dataset_details}

\begin{table}[!ht]
\centering
\caption{Comparison of zero parameter models to verify data integrity of Anidance benchmark.}
\begin{tabularx}{\columnwidth}{l@{\extracolsep{\fill}} ccc}
&\multicolumn{3}{c}{\textbf{L2P}} \\
\cmidrule(lr){2-4}
Length & 5 & 15 & 30 \\
\cmidrule(lr){1-4}
Zero Velocity~\cite{duan2021singleshot} & 2.34 & 5.12 & 6.73 \\
Zero Velocity (ours) & 2.44 & 5.15 & 6.89 \\
\cmidrule(lr){1-4}
LERP Interpolation~\cite{duan2021singleshot} & 0.94 & 3.24 & 4.68 \\
LERP Interpolation (ours)  & 0.94 & 3.06 & 4.84 \\

\bottomrule
\end{tabularx}%
\label{table:anidance_validation}
\end{table}%

The Anidance benchamark\footnote{Accesible at: \url{https://github.com/Music-to-dance-motion-synthesis/dataset}} was originally created for music-conditioned motion generation. \citet{duan2021singleshot} extracted the human motion data and formulated a dataset similar to LaFAN1 that contains only global joint coordinates.
In it's raw form the dataset is provided as a .json file. We took the following pre-processing steps:
\begin{enumerate}
    \item Convert the .json file into a Pandas dataframe with each column corresponding to a joint coordinate. Each joint has 3 coordinates $(x,y,z)$ and we have 24 joints in the Anidance skeleton so we have a total of 72 columns. Each row in the dataframe corresponds to one time step of the 25 FPS recording.
    \item Using the same data pipeline as with the LaFAN1 data we created windows of size 128 frames with an offset of 64 frames for all sequences. Note that unlike LaFAN1 we do not perform XZ-centering, rotation along the Y-axis or any other other pre-processing besides calculating the necessary statistics for the standardization of the L2P metric.
    \item After corresponding with~\citet{duan2021singleshot} we were able to replicate the data split. However, since the data pipeline for the code is not public we had to re-implement it from scratch. We recover the exact same number of frames (101,390) and number of test dance sequences (323) but we obtain slightly more training sequences (1172 vs. 1117) than reported by~\citet{duan2021singleshot}. The difference is smaller than 5\%.
    \item Our test data are identical, however since the L2P metric is standardized based on the training data this creates small discrepancies. To validate that the discrepancies are relatively minor we compare the L2P metric for zero-parameter deterministic models, namely zero-velocity and linear interpolation with respect to the reference values of~\citet{duan2021singleshot} in Table~\ref{table:anidance_validation}. As we can see, the L2P we recover for these baseline models is often identical with reported reference figures and in the worst case within 5\% and in general appear not to be biased in either the direction of overestimating or underestimating the error. Our results indicate improvements in the range of $\ge$30\% error reduction therefore our experimental setup is valid.
\end{enumerate}

\section{Additional Animations} 
A non-animated figure for comparison of tracer lines of the predicted joint positions by our model vs. SLERP and the ground truth is shown in Figure~\ref{fig:bonus_pic}. Additional demo videos\footnote{Accessible at:\url{https://storage.googleapis.com/delta-interpolator/additional-demo-videos.zip}} show long-horizon animation authoring using our approach with random-in-the-wild keyframe distributions outside the training set when we plug-in our model to the Unity editor. \textbf{For an objective evaluation of real world use cases of our model we strongly encourage viewing these videos.} Our model is able to consistently run at over 200 frames per second inside the Unity Editor.

\end{document}